\documentclass{article}

\usepackage{microtype}
\usepackage{graphicx}
\usepackage{subcaption}
\usepackage{booktabs} 
\usepackage{hyperref}

\usepackage[preprint]{icml2026}

\usepackage{amsmath}
\usepackage{amssymb}
\usepackage{mathtools}
\usepackage{amsthm}

\usepackage[capitalize,noabbrev]{cleveref}

\theoremstyle{plain}

\theoremstyle{definition}

\theoremstyle{remark}

\usepackage[textsize=tiny]{todonotes}

\usepackage{multirow}
\usepackage{bm}
\usepackage{makecell}
\usepackage{pifont}
\newcommand{\vect}[1]{\bm{#1}}
\icmltitlerunning{DyMoE: Dynamic Expert Orchestration with Mixed-Precision Quantization for Efficient MoE Inference on Edge}

\begin{document}

\twocolumn[
  \icmltitle{DyMoE: Dynamic Expert Orchestration with Mixed-Precision Quantization \\ for Efficient MoE Inference on Edge}



  \icmlsetsymbol{cor}{*}
  \begin{icmlauthorlist}
    \icmlauthor{Yuegui Huang}{sysu}
    \icmlauthor{Zhiyuan Fang}{sysu,cor}
    \icmlauthor{Weiqi Luo}{sysu}
    \icmlauthor{Ruoyu Wu}{tencent}
    \icmlauthor{Wuhui Chen}{sysu}
    \icmlauthor{Zibin Zheng}{sysu}

  \end{icmlauthorlist}

  \icmlaffiliation{sysu}{Sun Yat-sen University, Guangzhou, China}
  \icmlaffiliation{tencent}{Tencent, Shenzhen, China}
  \icmlcorrespondingauthor{Zhiyuan Fang}{fangzhy27@mail2.sysu.edu.cn}

\icmlkeywords{Large Language Models, Mixture-of-Experts, Model Quantization, Offloading}
  \vskip 0.3in
]



\printAffiliationsAndNotice{}  

\begin{abstract}
Despite the computational efficiency of MoE models, the excessive memory footprint and I/O overhead inherent in multi-expert architectures pose formidable challenges for real-time inference on resource-constrained edge platforms. While existing static methods struggle with a rigid latency-accuracy trade-off, we observe that expert importance is highly skewed and depth-dependent. Motivated by these insights, we propose DyMoE, a dynamic mixed-precision quantization framework designed for high-performance edge inference. Leveraging insights into expert importance skewness and depth-dependent sensitivity, DyMoE introduces: (1) importance-aware prioritization to dynamically quantize experts at runtime; (2) depth-adaptive scheduling to preserve semantic integrity in critical layers; and (3) look-ahead prefetching to overlap I/O stalls. Experimental results on commercial edge hardware show that DyMoE reduces Time-to-First-Token (TTFT) by $3.44\times$\text{--}22.7$\times$ and up to a $14.58\times$ speedup in Time-Per-Output-Token (TPOT) compared to state-of-the-art offloading baselines, enabling real-time, accuracy-preserving MoE inference on resource-constrained edge devices.
\end{abstract}

\section{Introduction}

Large Language Models (LLMs) are increasingly transitioning from centralized cloud services to local deployment on resource-constrained edge platforms, driven by imperatives such as data privacy, zero-latency availability, and inference cost sustainability~\cite{yu2024edge,zheng2025review, xu2024device}. Simultaneously, the Mixture-of-Experts (MoE)~\cite{shazeer2017outrageously} architecture has emerged as the dominant paradigm for scaling LLMs by leveraging sparse activation to decouple model capacity from computational cost. By dynamically routing tokens to a sparse subset of experts, MoE delivers massive-model reasoning capabilities with significantly lower FLOPs than dense models of equivalent scale. Although originally designed for cloud-scale efficiency, this sparse, low-compute characteristic incidentally makes high-performance inference theoretically feasible on hardware-limited edge devices.

\begin{figure}
    \centering
    \includegraphics[width=1\linewidth]{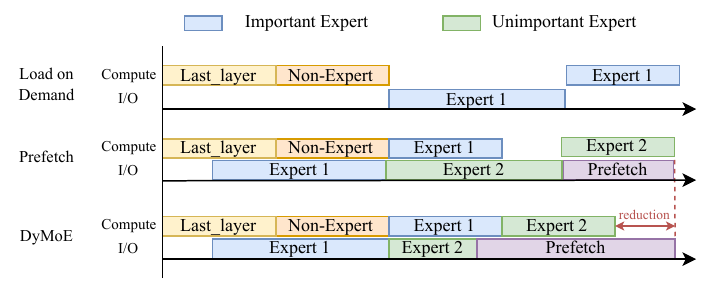}
    \caption{Pipeline Comparison: DyMoE vs. Two Conventional MoE Baselines.}
    \label{fig:pipeline}
\end{figure}

However, despite their computational efficiency, deploying high-performance MoE models on edge devices faces a daunting storage and bandwidth wall. The massive inactive parameter set creates a footprint that far exceeds the physical memory of typical edge hardware; for instance, Mixtral-8$\times$7B requires approximately 87~GB in BF16 format, whereas consumer laptops or embedded AI accelerators often possess significantly less memory. To mitigate this, offloading inactive experts to host memory is a primary strategy. However, as illustrated in~\autoref{fig:pipeline}, a naive load-on-demand approach incurs prohibitive latency. Even with prefetching~\cite{yi2025edgemoe,tang2024hobbit,zhou2025floe,fang2025fate}, the system still suffers from substantial I/O bubbles, as the time to load massive expert weights typically dwarfs the narrow computation window, leaving the GPU idle in Wait-for-Weight stalls.

To alleviate this bandwidth pressure, existing optimizations focus on reducing data volume through weight compression or expert skipping. However, these methods often suffer from a lack of fine-grained adaptivity to both the model's intrinsic structure and the dynamic nature of inference. Uniform quantization~\cite{frantar2022gptq,lin2024awq,badri2023hqq,xiao2023smoothquant} treats all parameters with equal importance, disregarding their varying sensitivity across layers and leading to disproportionate accuracy loss at extreme bit-widths (e.g., Int2). Meanwhile, static mixed-precision~\cite{yi2025edgemoe, zhou2025floe} and dynamic skipping frameworks~\cite{zhong2024adapmoe, tang2024hobbit,lu2024not} remain constrained by offline-derived statistics or pre-determined thresholds. Because these strategies are frozen prior to deployment, they are incapable of adapting to the fluid dynamics of real-time input. 

The limitations of static approaches motivate a more granular exploration of MoE inference dynamics. In this work, we identify three synergistic properties that bridge algorithmic sparsity with potential system-level efficiency: (1) Dynamic Skewness: expert importance is primarily driven by a small subset of highly influential tokens—often referred to as heavy-hitters—whose activation patterns vary significantly across inputs, suggesting that uniform expert treatment is inherently sub-optimal; (2) Depth-Dependent Sensitivity: model layers exhibit non-uniform tolerance to information loss, with deeper layers demonstrating significantly higher quantization robustness; and (3) Inter-layer Predictability: the inherent activation similarity across adjacent layers enables the accurate look-ahead identification of critical experts for subsequent stages.

Guided by these insights, we propose DyMoE, an algorithm-system co-designed MoE inference framework that requires zero re-training or calibration overhead. DyMoE introduces a dynamic precision scheduler that assigns experts to a spectrum of mixed-precision states (e.g., 8-bit, 4-bit, and ``0-bit'') based on their runtime importance. Under this unified representation, a 0-bit assignment corresponds to expert skipping, effectively eliminating both memory I/O overhead and computational costs for redundant parameters. To mitigate the I/O bottlenecks inherent in edge devices, DyMoE leverages inter-layer predictability to implement a look-ahead prefetching mechanism, which overlaps expert loading with active computation. As illustrated in \autoref{fig:pipeline}, by dynamically modulating the precision of expert weights, DyMoE significantly accelerates inference throughput on resource-constrained commodity hardware while maintaining competitive model performance.

Our specific contributions are as follows:

\begin{itemize} \item \textbf{Dynamic Expert Importance Classification Method:} We propose a runtime-adaptive scheme that classifies experts into Critical and Sub-critical tiers based on heavy-hitter tokens and gate score. By incorporating depth-dependent sensitivity, our method ensures that limited resources are strictly prioritized for experts most vital to model performance.

\item \textbf{A High-Performance Inference System:} 
Leveraging the proposed classification, we implement a Dynamic Mixed-Precision Expert Orchestration system that dynamically assigns experts to optimal execution paths. Specifically, sub-critical experts are transitioned into lower-precision quantization to minimize resource consumption. This system is powered by two synergistic components:
(i) a \textbf{Look-ahead Prefetching Engine} that exploits inter-layer predictability to prefetch critical weights, effectively overlapping I/O with computation; and 
(ii) \textbf{Mixed-precision Cache Management} that orchestrates limited VRAM.

\item \textbf{Extensive Empirical Validation:} 
We evaluate DyMoE on representative MoE architectures (e.g., Mixtral-8$\times$7B~\cite{jiang2024mixtral}, Qwen3-30B-A3B~\cite{qwen3}) across diverse edge-side memory constraints (12\text{--}24~GB). Experimental results demonstrate that DyMoE achieves a $3.44\times$\text{--}22.7$\times$  reduction in TTFT and up to a $14.58\times$ speedup in Time-Per-Output-Token latency (TPOT) compared to state-of-the-art offloading baselines. Crucially, these gains are achieved with marginal accuracy degradation.
\end{itemize}

\section{Background and Related Work}

\subsection{Mixture-of-Experts Architecture} \label{sec:background}

\begin{figure}[t]
    \centering
    \begin{subfigure}[b]{0.33\linewidth}
        \centering
        \includegraphics[width=\linewidth]{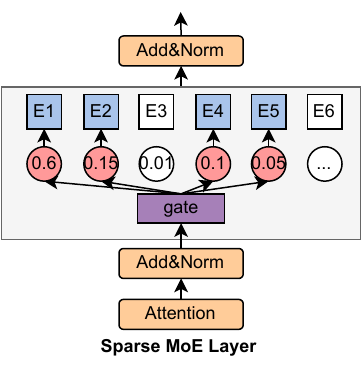}
        \caption{MoE Architecture}
        \label{fig:MoE} 
    \end{subfigure}
    \hfill 
    \begin{subfigure}[b]{0.64\linewidth}
        \centering
        \includegraphics[width=\linewidth]{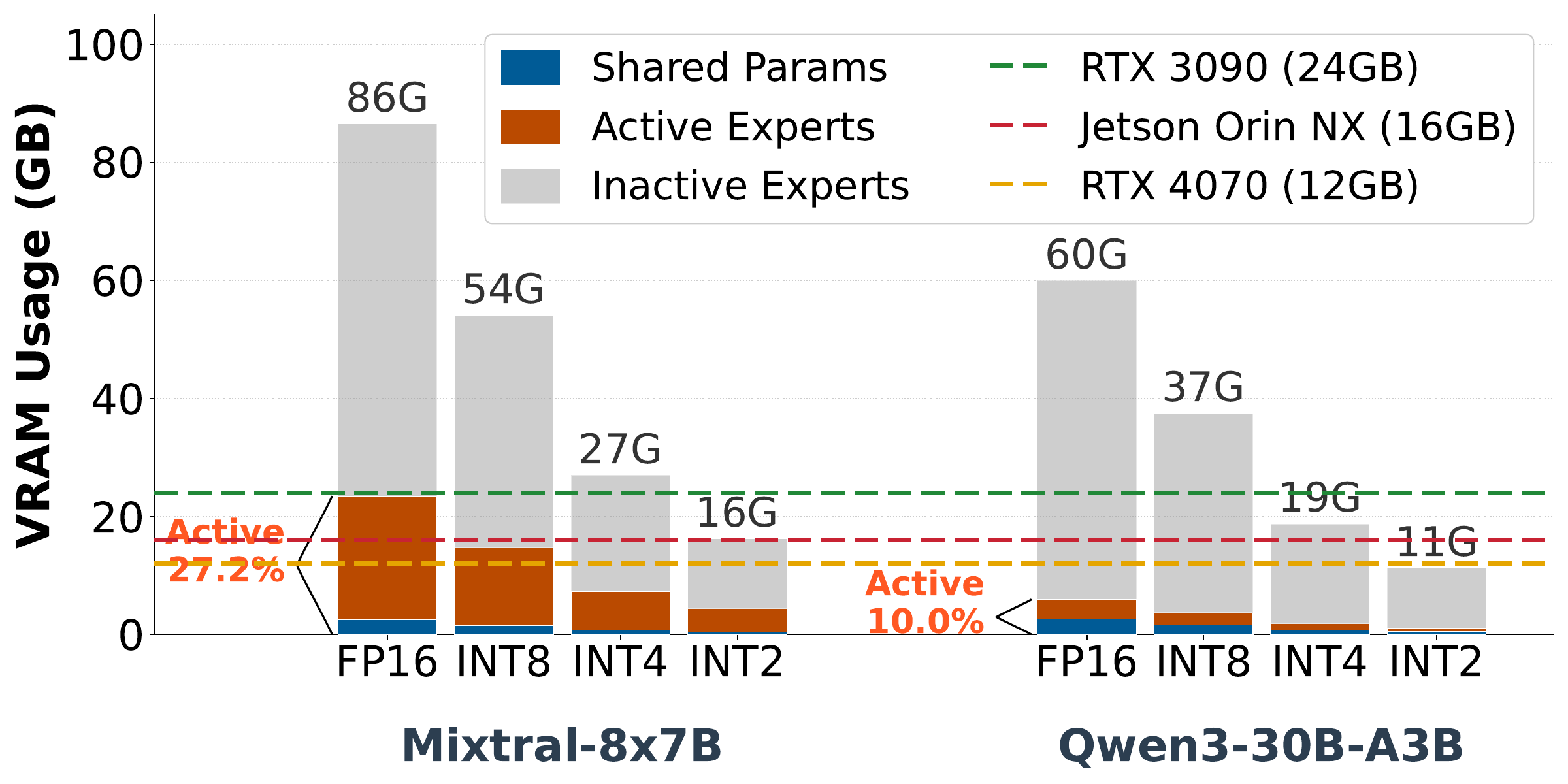}
        \caption{Memory Demands of SOTA MoEs}
        \label{fig:moe_memory}
    \end{subfigure}
    
    \caption{Overview of MoE Structure and its Memory Demands.}
    \label{fig:moe_background}
\end{figure}

The Mixture-of-Experts (MoE)\cite{shazeer2017outrageously} architecture scales model capacity by replacing dense Feed-Forward Networks (FFNs) with a sparse layer comprising multiple independent experts and a gating network (~\autoref{fig:MoE}). During inference, the router selects a small subset of experts per token, effectively maintaining a constant computational cost (FLOPs) despite a massive parameter footprint. 

However, this architectural efficiency in computation does not translate to memory savings. As illustrated in~\autoref{fig:moe_memory}, modern MoE models, such as Mixtral-$8\times7B$ and Qwen3-30B-A3B, possess parameter footprints that far exceed the VRAM capacity of common edge hardware (e.g., 12~GB, 16~GB, or 24~GB). Paradoxically, while these models require prohibitive storage, their runtime utilization is remarkably sparse: Mixtral-$8\times7B$ activates only $\sim$27\% of its parameters per token, while Qwen3-30B-A3B activates as little as 10\%. This vast memory-utilization gap---where 70--90\% of parameters remain idle at any given step---motivates our pursuit of a runtime-adaptive system.

\subsection{MoE Compression and Offloading}
To alleviate the severe memory bottlenecks, researchers have explored various model compression methods and system-level optimization techniques.

\noindent\textbf{Quantization and Pruning.} Standard Post-Training Quantization (PTQ) methods like GPTQ~\cite{frantar2022gptq} and AWQ~\cite{lin2024awq} apply uniform bit-widths to reduce model footprint. However, as illustrated in~\autoref{fig:moe_memory}, even uniform 4-bit quantization is often insufficient to fit large MoE models into limited edge VRAM; meanwhile, aggressive 2-bit quantization typically leads to catastrophic accuracy collapse (~\autoref{tab:model_comparison_slim}). While static mixed-precision (e.g., EdgeMoE~\cite{yi2025edgemoe}) addresses this, it remains oblivious to dynamic input complexity at runtime. Alternatively, structural pruning~\cite{lu2024not} or expert merging~\cite{li2023merge} reduces model size but incurs irreversible information loss and prohibitive re-training overhead.

\noindent\textbf{System-Level Offloading.} Parameter offloading utilizes host RAM to bypass GPU memory limits. However, throughput-oriented frameworks~\cite{, holmes2024deepspeed, cao2025moe,fang2025klotski} rely on large batch sizes to hide I/O latency, making them ineffective for latency-sensitive edge scenarios where the batch size is one. Heterogeneous strategies like Fiddler~\cite{kamahori2024fiddler} offload certain computations to the CPU but encounter compute-bound bottlenecks during dequantization, leading to latency penalties that outweigh transmission savings. In summary, a gap exists for a system that can orchestrate mixed-precision and execution paths dynamically. 

\begin{figure}[t]
    \centering
    \includegraphics[width=1\linewidth]{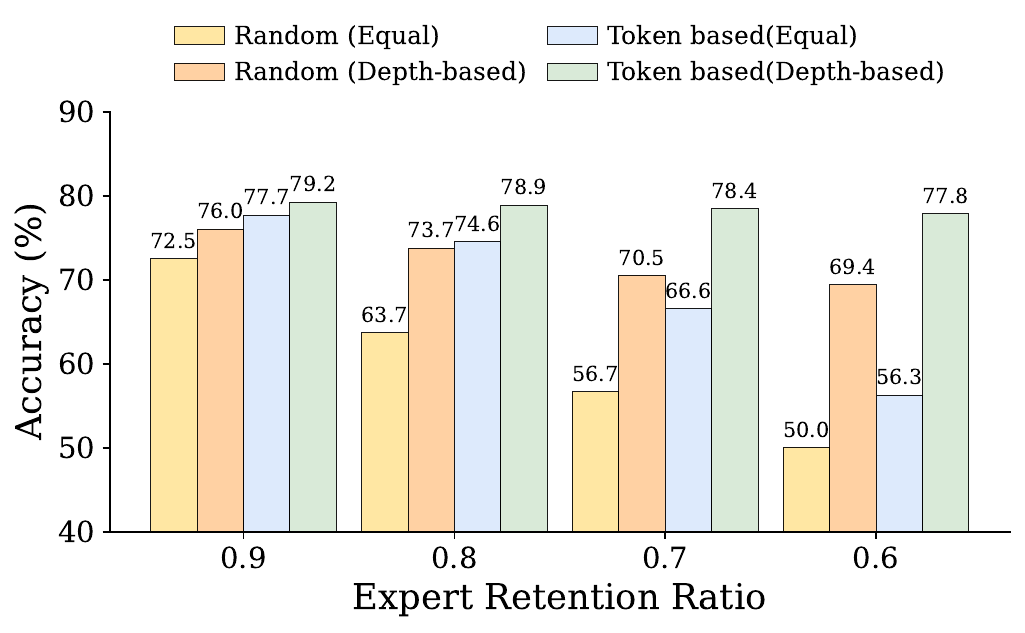}
    \caption{
        Performance evaluation of various expert pruning strategies on the C-Eval benchmark across different retention ratios.
        \textit{Random}: experts are retained randomly; 
        \textit{Token-based}: experts are prioritized based on the volume of assigned critical tokens; 
        \textit{Equal}: applies a uniform pruning ratio across all layers; 
        \textit{Depth-based}: adjusts the retention ratio dynamically according to layer depth.
    }
    \label{fig:pruning_comparison}
\end{figure}

\section{Observations}
\label{sec:observations}

\subsection{Dynamic Skewness in Expert Importance}
\label{subsec:obs_importance}

\noindent\textbf{Quantifying Expert Importance via Token Load.} While prior research~\cite{yang2024pyramidinfer,zhang2023h2o,zhou2025sparseserve} has established that LLM performance is primarily sustained by a sparse subset of heavy-hitter tokens, we posit that this non-uniformity naturally extends to the expert level through the MoE routing mechanism. Specifically, we quantify an expert’s functional contribution by the volume of critical tokens it processes. Our expert skipping experiments, illustrated in \autoref{fig:pruning_comparison}, validate this importance-driven selection logic. By prioritizing experts that process a higher volume of critical tokens, we consistently maintain model performance across diverse expert retention ratios.

\begin{figure}[t]
    \centering
    \includegraphics[width=1\linewidth]{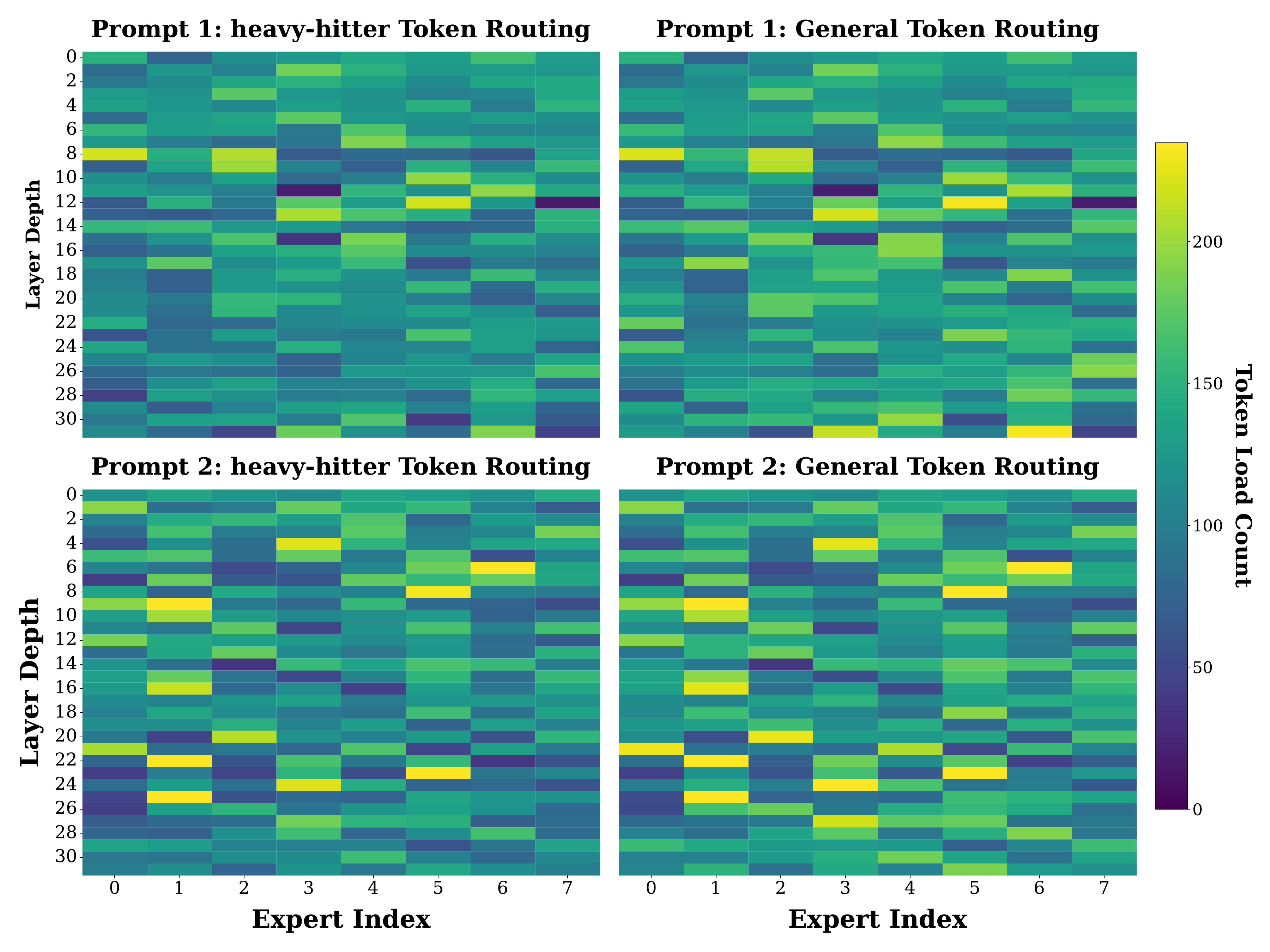}
    \caption{Comparative Visualization of Expert Routing Distributions: heavy-hitter and general Tokens across Different Inputs.}
    \label{fig:Token_load}
\end{figure}

Analyzing the distribution of these critical tokens across experts (as visualized in~\autoref{fig:Token_load}) reveals that critical tokens are not uniformly scattered but cluster on specific semantic hotspots that shift dynamically across different inputs, invalidating static prioritization methods.

\subsection{Depth-Aware Sensitivity to Quantization}
\label{subsec:obs_sensitivity}

\begin{figure}[t]
\centering
\includegraphics[width=1 \linewidth]{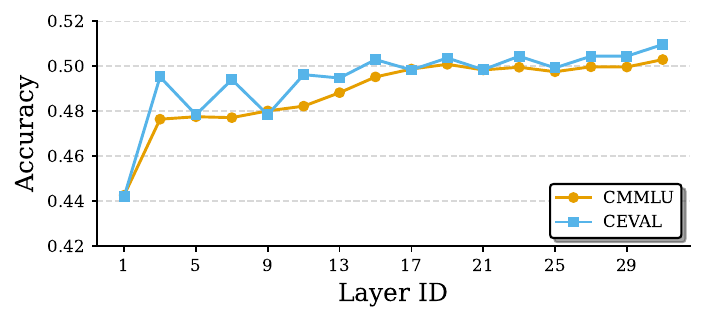}
\caption{Layer-wise sensitivity of Mixtral-8x7B-Instruct under Int2 quantization, measured on C-Eval and CMMLU benchmarks.}
\label{fig:layerwise_experiment}
\end{figure}

To determine the optimal precision allocation, we investigate how quantization robustness evolves with network depth. We perform a sensitivity analysis on Mixtral-8$\times$7B-Instruct by independently quantizing the experts of a single layer to Int2 while maintaining all other layers in BF16.
As illustrated in~\autoref{fig:layerwise_experiment}, the model exhibits a distinct depth-dependent sensitivity pattern. Shallow layers are exceptionally intolerant to quantization noise, where aggressive Int2 quantization precipitates a pronounced decline in accuracy; in sharp contrast, deeper layers demonstrate remarkable resilience, tolerating such extreme compression.

This sensitivity pattern is twofold. First, shallow layers perform fundamental feature extraction that is critical for all subsequent computations. Second, precision loss in these early stages undergoes cumulative amplification as it propagates through the network, leading to catastrophic downstream errors. In contrast, deeper layers exhibit significantly higher noise resilience, permitting more aggressive quantization without compromising semantic integrity.

\begin{figure}[t]
\centering
\includegraphics[width=1 \linewidth]{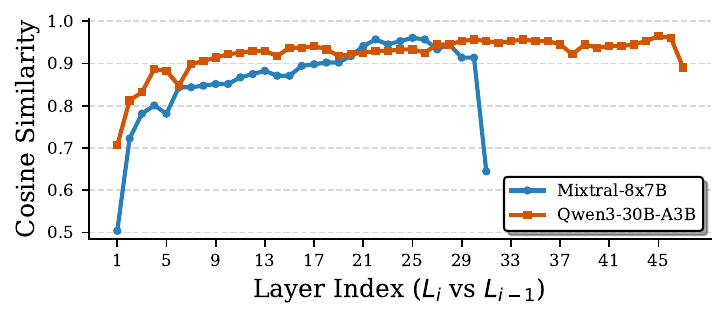}
\caption{Adjacent Layer Cosine Similarity.}
\label{fig:obs_similarity}
\end{figure}
\subsection{Inter-layer Activation Similarity and Look-ahead Opportunities}
\label{subsec:obs_similarity}

Transformer architectures rely heavily on residual connections, leading to high semantic stability across adjacent layers.
To quantify this, we analyze the cosine similarity of activations between consecutive layers. As illustrated in~\autoref{fig:obs_similarity}, both Qwen3-30B-A3B and Mixtral-8$\times$7B exhibit consistently high similarity scores across the network depth.
This empirical evidence confirms that the hidden state $\bm{h}^{(l)}$ serves as a high-fidelity proxy for $\bm{h}^{(l+1)}$.

Prior research~\cite{eliseev2023fast,tang2024hobbit} has leveraged inter-layer similarity to predict expert activation for the subsequent layer. However, our framework requires a more granular distinction: identifying critical experts rather than mere activation. As demonstrated in~\autoref{fig:Token_load}, there is a high statistical correlation between an expert's total token load and its heavy-hitter token load, suggesting that the token distribution serves as a robust proxy for the importance distribution. By leveraging this proxy, we can proactively forecast the importance of experts, enabling the system to orchestrate their fidelity states.

\textbf{Summary.} In short, these empirical observations provide a three-fold foundation for our design: (1) the skewed importance enables the dynamic identification of expert importance at runtime; (2) the depth-dependent sensitivity allows for aggressive compression in deeper layers; and (3) the inter-layer similarity offers a window for prefetching. Together, these insights motivate the architecture of DyMoE.

\begin{figure}[t]
\centering
\includegraphics[width=1 \linewidth]{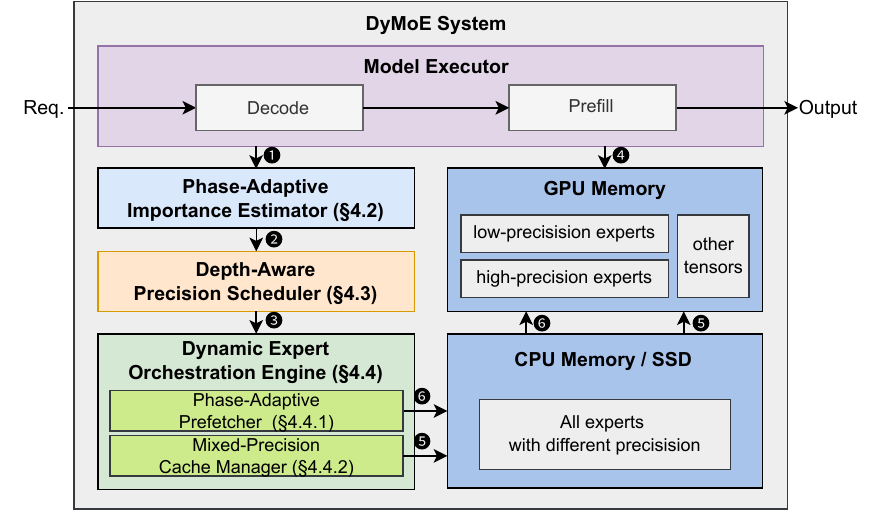}
\caption{Overview of DyMoE.}
\label{fig:overview}
\end{figure}

\section{DyMoE Design}
\label{sec:DyMoEDesign}

\subsection{System Overview}
Building upon our empirical insights, we propose DyMoE, an algorithm-system co-designed framework that transforms static MoE execution into a dynamic, mixed-precision inference.
\autoref{fig:overview} illustrates the architectural orchestration of DyMoE. The workflow initiates with the Phase-Adaptive Expert Importance Estimator, which assesses expert significance via token-level metrics (\ding{182}). Based on these scores, the Depth-Aware Precision Scheduling determines the precision allocation for the Dynamic Expert Orchestration Engine (\ding{183}). As a core constituent of the Dynamic Expert Orchestration Engine, Mixed-Precision Cache Manager dispatches hit experts stored in VRAM to the execution pipeline upon receiving the scheduling decision (\ding{184}). Simultaneously, it triggers asynchronous requests to fetch absent experts from CPU Memory or SSD at their designated precision, ensuring that the Model Executor operates on a unified mixed-precision weight set (\ding{185}). Crucially, DyMoE leverages a Phase-Adaptive Prefetcher to predict and pre-load critical experts based on intermediate activations, effectively masking I/O latency by overlapping weight loading with ongoing expert computation and non-MoE operations (\ding{186}, \ding{187}).

\subsection{Phase-Adaptive Expert Importance Estimator}
\label{subsec:phase_selection}

Recognizing the divergent characteristics of MoE inference across execution stages, DyMoE adopts tailored strategies to estimate expert importance for the prefill and decoding.

\subsubsection{Prefill: Token-Guided Importance}
To operationalize the dynamic skewness observed in~\autoref{subsec:obs_importance}, DyMoE implements a token-guided importance scoring mechanism tailored for the prefill phase. Unlike the decoding phase, the prefill stage provides access to the complete input sequence, enabling a global assessment of token-level significance before the experts are invoked.

We first quantify the semantic importance $s_i$ of each token $t_i$ by aggregating attention weights across all heads, capturing the token's influence on the overall sequence context:
\begin{equation}
s_i = \frac{1}{H} \sum_{h=1}^{H} a_{i}^{(h)},
\end{equation}
where $H$ denotes the number of attention heads and 
$a_{i}^{(h)}$ represents the attention score of token $t_i$ in the $h$-th head.

\begin{figure}[t]
\centering
\includegraphics[width=1\linewidth]{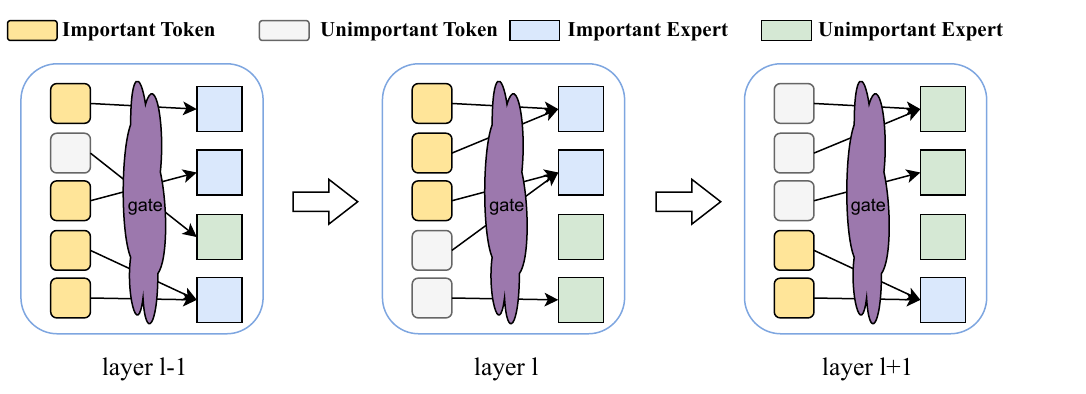}
\caption{Token-Guided Expert Selection (Prefill). Experts processing more heavy-hitter tokens are prioritized.}
\label{fig:Prefill_figure}
\end{figure}

Building on the insight that an expert's utility is an inheritance of the tokens it processes, we define the importance of expert $E_j$ as its heavy-hitter token load (as shown in~\autoref{fig:Prefill_figure}). Let $\mathcal{T}_{\text{imp}}$ be the set of top-$k$ tokens with the highest $s_i$ scores. The importance metric for expert $E_j$ is calculated as the number of critical tokens routed to it:
\begin{equation}
\mathcal{I}_{\text{Prefill}}(E_j) = \left| \{ t_i \in \text{Tokens}_j \mid t_i \in \mathcal{T}_{\text{imp}} \} \right|,
\end{equation}
where $\text{Tokens}_j$ is the set of tokens assigned to $E_j$. This metric serves as the decision criterion for DyMoE's mixed-fidelity orchestration, enabling the system to dynamically assign optimal execution policies for every expert.

\subsubsection{Decode: Gate-Guided Importance}
In the decode phase, tokens are generated sequentially, rendering statistical aggregation ineffective. Therefore, we rely on the gating mechanism itself to estimate expert importance. We define expert importance directly using the gating score vector $\vect{g}$, where $g_j$ represents the routing weight assigned to expert $E_j$. Since the gating score dictates the contribution weight of an expert to the final output, it serves as a precise proxy for runtime importance:
\begin{equation}
\mathcal{I}_{\text{Decode}}(E_j) = g_j.
\end{equation}
As shown in~\autoref{fig:Decode_figure}, experts with higher gating scores are selected as critical, ensuring that the model's most confident routing paths are executed in high precision.

\subsection{Depth-Aware Precision Scheduling}
\label{sec:depth_control}
\begin{figure}
\centering
\includegraphics[width=1\linewidth]{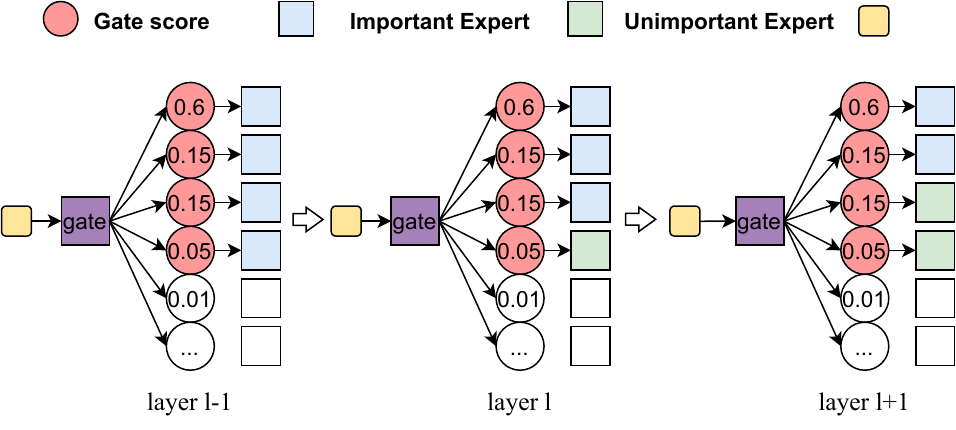}
\caption{Gate-Guided Expert Selection (Decode). Experts with higher routing scores are selected as critical.}
\label{fig:Decode_figure}
\end{figure}

Based on the sensitivity observations in \autoref{subsec:obs_sensitivity}, we propose a depth-aware scheduler that allocates more high-precision experts to early layers (shown in~\autoref{fig:Prefill_figure} and~\autoref{fig:Decode_figure}). Specifically, we define the retention ratio $r(l)$ at layer $l$ using a cosine schedule:
\begin{equation}
r(l) = (1-\lambda)\cdot \frac{\cos\left(\pi \cdot \frac{l}{L-1}\right)+1}{2} + \lambda,
\end{equation}
where $\lambda \in [0,1]$ is a hyperparameter that controls the overall retention ratio of all experts.

Therefore, the number of critical experts $t_l$ is calculated as:
\begin{equation}
\label{eq:t_l}t_l = \left\lceil r(l) \cdot M \right\rceil.
\end{equation}

We choose the cosine function because it stays near 1 at the beginning layers and then decreases smoothly. Unlike a linear decay that drops immediately, this "slow-start" characteristic ensures that sensitive early-stage features are better preserved. As depth increases, it provides a graceful transition to a lower budget in the deeper, more robust layers.

\subsection{Dynamic Expert Orchestration Engine}
\label{sub_sec:system_impl}

To fully exploit expert heterogeneity, DyMoE implements a dynamic orchestration engine. This engine leverages expert heterogeneity through two intertwined pillars: Phase-Adaptive Prefetcher, which proactively overlaps critical expert transfers with computation, and a mixed-precision expert cache, which maximizes expert residency under tight VRAM constraints by storing experts at different precisions based on their importance.

\subsubsection{Phase-Adaptive Prefetcher}
To mask the latency of loading critical experts, we must identify them before the computation of the current layer is complete. Exploiting the inter-layer activation similarity (~\autoref{subsec:obs_similarity}), we approximate the gating scores for the next layer $l+1$ using the current hidden state $\bm{h}^{(l)}$:
\begin{equation}
\hat{\vect{g}}^{(l+1)}_i = \operatorname{Softmax}(\bm{h}^{(l)}_i \bm{W}_g^{(l+1)}).
\label{eq:gating_approx}
\end{equation}
Based on these approximated scores, we predict the likely-to-be-activated experts $\vect{s}_i^{(l)} = \operatorname*{TopK}_k(\hat{\vect{g}}^{(l+1)}_i)$.

 We employ different prefetching strategies for Prefill and Decode phases:

\noindent\textbf{Prefill (Token-Frequency Prefetching).} 
Since the exact token-level demand fluctuates in parallel processing, we aggregate the predicted demand across all tokens in the batch. We calculate the activation frequency $c_e$ for each expert $e$:
\begin{equation}
c_e = \sum_{i \in \mathcal{T}} \mathbb{I}\!\left( e \in \vect{s}_i^{(l)} \right).
\label{eq:freq}
\end{equation}
We then prefetch the top-$t$ experts with the highest frequency $c_e$ into the GPU cache, maximizing the hit rate for the batch.

\noindent\textbf{Decode (Direct Prefetching).} 
For sequential decoding, predictions are specific to the single current token. We directly prefetch the top-$t$ predicted experts:
\begin{equation}
\mathcal{P}^{(l)}_{\text{dec}} = \operatorname*{TopK}_t\!\left( \hat{\vect{g}}^{(l+1)}_i \right).
\label{eq:Decode_prefetch}
\end{equation}

\subsubsection{Mixed-Precision Cache Management}
To manage limited VRAM efficiently, we extend the standard LRU cache to support mixed-precision storage, governed by three rules:
\begin{itemize}
    \item \textbf{No Duplication:} An expert is stored in only one format (High or Low) to prevent redundancy.
    \item \textbf{Precision Promotion:} If a request requires High Precision but only Low Precision is cached, the system treats it as a miss, loading the High-Precision Experts from host memory and evicting the Low-Precision version.
    \item \textbf{Conservative Reuse:} If Low Precision is requested but High Precision is cached, the High-Precision version is reused to maintain accuracy without additional I/O.
\end{itemize}

\section{Implementation}
Given that MoE experts account for the vast majority of model parameters, we focus quantization exclusively on these layers to maximize memory efficiency. In our implementation, we employ GPTQ~\cite{frantar2022gptq} as the basic quantization algorithm due to its robust post-training performance. It is important to note that DyMoE can be seamlessly integrated with other advanced quantization techniques, such as AWQ~\cite{lin2024awq} and HQQ~\cite{badri2023hqq}. We evaluate two primary configurations on DyMoE:  "$4/2$", which retains critical experts in 4-bit while compressing sub-critical ones to 2-bit, and "$4/0$", which entirely bypasses non-critical experts.

\section{Evaluation}
\label{sec:evaluation}

\subsection{Experimental Setup}
\label{subsec:setup}

\noindent\textbf{Hardware.} 
Experiments are conducted on a server with an AMD EPYC 7542 CPU and an NVIDIA RTX 3090 GPU (24~GB) via PCIe Gen3 $\times$16. To simulate resource-constrained edge environments (12--24~GB), we employ a software-level memory allocator to strictly limit VRAM.

\noindent\textbf{Models.} 
We evaluate DyMoE on Mixtral-8$\times$7B and Qwen3-30B-A3B, representing both coarse-grained (low-sparsity) and fine-grained (high-sparsity) MoE architectures.


\noindent\textbf{Baselines.} 
We compare DyMoE with four SOTA inference systems to evaluate its efficiency. 
(1) Accelerate~\cite{accelerate}, a widely-used framework supporting heterogeneous device partitioning and quantization integration. 
(2) Mixtral-Offloading~\cite{eliseev2023fast}, an MoE-specific framework utilizing LRU expert caching and mixed-precision support. 
(3) MoE-Infinity~\cite{xue2024moe}, a system employing activation-aware prefetching and fine-grained caching strategies. 
(4) Fiddler~\cite{kamahori2024fiddler}, a CPU--GPU co-execution framework that dynamically offloads computation to relieve GPU bottlenecks. 

\noindent\textbf{Workload.} 
To faithfully emulate real-world usage patterns, we evaluate end-to-end latency using input sequences sampled from the ShareGPT~\cite{sharegpt2023} dataset. To simulate latency-sensitive edge deployment, all experiments are conducted with a batch size of 1, representing continuous single-user serving.
Complementarily, we assess model accuracy using the LM Evaluation Harness~\cite{eval-harness} across multiple benchmarks, including MMLU~\cite{hendryckstest2021}, CMMLU~\cite{li2023cmmlu}, and GSM8K~\cite{cobbe2021gsm8k}.

\noindent\textbf{Key Metrics.} 
We focus on two critical latency metrics for interactive applications: Time-to-First-Token (TTFT) and Time-Per-Output-Token (TPOT). Additionally, we report the accuracy on the aforementioned benchmarks to evaluate the impact of our Dynamic Quantization scheme.

\subsection{End-to-End System Efficiency}
\begin{figure}[t]
\centering
\includegraphics[width=1\linewidth]{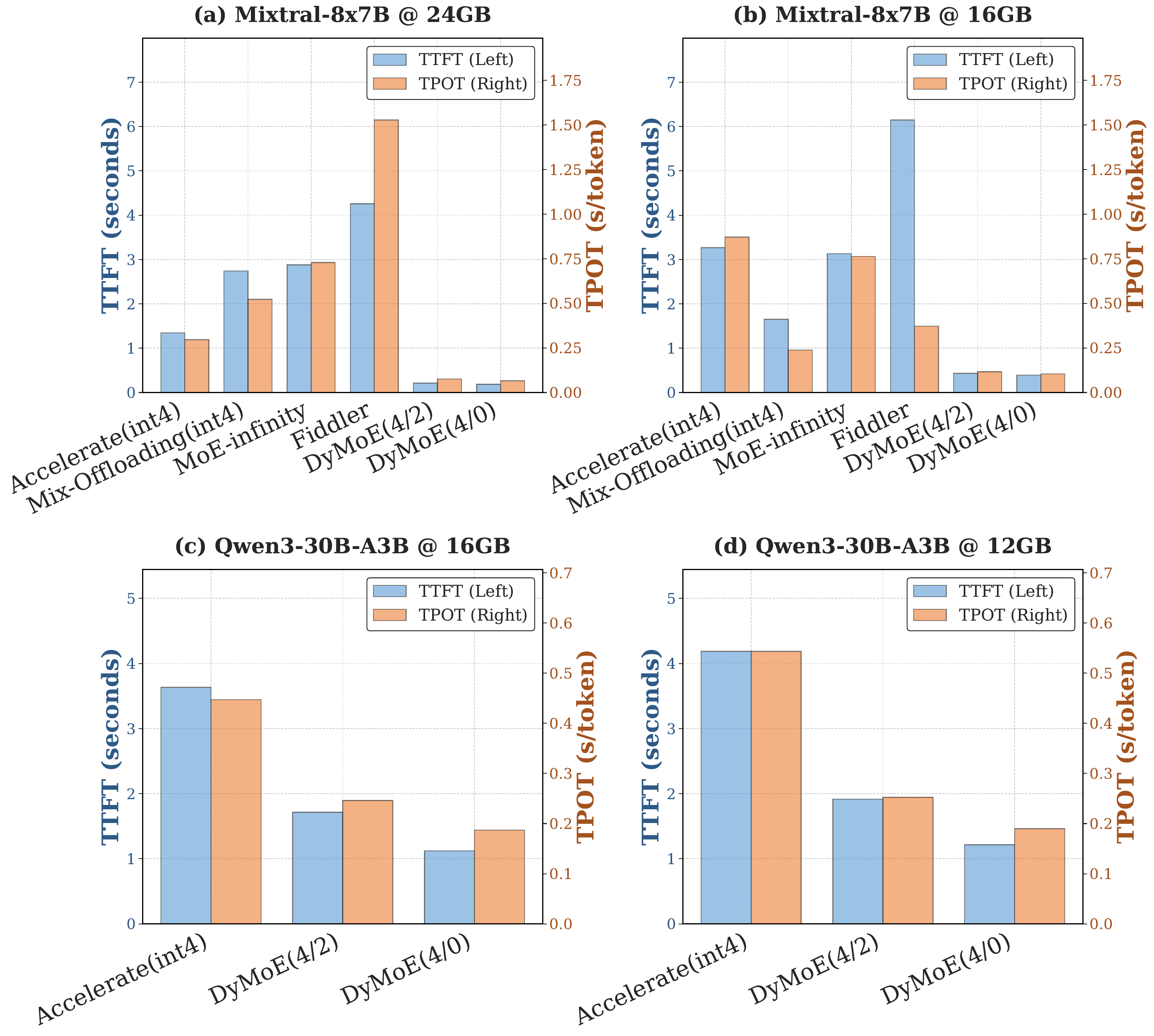}
\caption{End-to-End Performance Comparison.}
\label{fig:performance}
\end{figure}

We evaluate the end-to-end efficiency of DyMoE using a default expert retention ratio of $r=0.75$, a setting that our accuracy benchmarks confirm to have a marginal impact on Model accuracy. As illustrated in \autoref{fig:performance}, DyMoE consistently outperforms all baselines across diverse hardware configurations and model scales.

\noindent\textbf{Prefill Latency (TTFT).}Under identical hardware environments, DyMoE demonstrates a dominant lead in prefill speed. For Mixtral-8x7B, DyMoE (4/0) achieves a peak speedup of 22.7$\times$ over Fiddler in the 24~GB setting and 15.7$\times$ in the 16~GB setting. When compared to quantization-based offloading frameworks, DyMoE maintains a significant edge, outperforming Accelerate by 7.2$\times$--8.3$\times$ and Mixtral-Offloading by 4.2$\times$--14.58$\times$ under the same memory constraints. This efficiency extends to high-sparsity models like Qwen3-30B-A3B, where DyMoE reduces TTFT by 3.44$\times$ in a 12~GB environment compared to the Accelerate baseline.
We attribute these gains to two primary factors: (i) our importance-aware scheduling, which identifies and quantizes sub-critical experts to low precision, thereby drastically reducing the total I/O volume; and (ii) our cache design, which maximizes the hit rate for experts within the constrained VRAM. By applying extreme quantization to sub-critical experts, the system alleviates the I/O bottleneck, effectively minimizing latency inducing host-to-device transfers over the PCIe bus.

\noindent\textbf{Decode Latency (TPOT).}DyMoE similarly excels in the decoding phase through runtime-aware expert orchestration. For Mixtral-8x7B in a 16~GB environment, it achieves a 14.58$\times$ reduction in per-token latency compared to Fiddler. Against quantized baselines in their respective identical environments, DyMoE yields a 8.31$\times$ speedup over Accelerate and up to a 2.27$\times$ speedup over Mixtral-Offloading. For the high-sparsity Qwen3-30B-A3B, DyMoE achieves a 2.86$\times$ decoding speedup in the same 12~GB environment, proving its robustness across diverse MoE architectures.The primary challenge in decoding is the serialized Wait-for-Weight stall. DyMoE mitigates this by maximizing the overlap between I/O and computation. Our look-ahead prefetching engine, powered by the inter-layer similarity observations in \autoref{sec:observations}, allows the system to fetch required critical experts into the GPU cache before they are needed for computation. Simultaneously, because mixed-precision quantization reduces the sheer amount of data to be transferred, the time required for I/O is often fully masked by the computation of non-MoE layers (e.g., Attention).

\begin{table}[t]
\centering
\caption{Accuracy Comparison under Uniform Quantization.}
\label{tab:model_comparison_slim}
\resizebox{0.85\linewidth}{!}{%
\begin{tabular}{@{}llccc@{}}
\toprule
\textbf{Dataset} & \textbf{Model} & \textbf{Int2} & \textbf{Int4} & \textbf{BF16} \\ \midrule
\multirow{2}{*}{\textbf{MMLU}}   & Mixtral-8$\times$7B  & 0.4979 & 0.6795 & 0.6889 \\
                                 & Qwen3-30B-A3B        & 0.4950 & 0.7705 & 0.7800 \\ \midrule
\multirow{2}{*}{\textbf{CMMLU}}  & Mixtral-8$\times$7B  & 0.3048 & 0.5044 & 0.5108 \\
                                 & Qwen3-30B-A3B        & 0.4300 & 0.8044 & 0.8063 \\ \midrule
\multirow{2}{*}{\textbf{GSM8K}}  & Mixtral-8$\times$7B  & 0.0781 & 0.6467 & 0.6467 \\
                                 & Qwen3-30B-A3B        & 0.5292 & 0.8908 & 0.8954 \\ \bottomrule
\end{tabular}%
}
\end{table}

\subsection{Inference Accuracy}

Having established the substantial efficiency gains, we next turn to the critical question of model integrity. In this section, we rigorously evaluate the impact of dynamic mixed-precision quantization on the model's generative capabilities. Our analysis demonstrates that DyMoE achieves these accelerations with negligible accuracy degradation, effectively safeguarding the model's original reasoning power.

\noindent\textbf{Robustness to Dynamic Quantization}
The experimental results (~\autoref{tab:combined_results}) demonstrate that DyMoE exhibits remarkable robustness across different model architectures. 
For Mixtral-8$\times$7B, the "4/0" configuration with $r=0.9$ achieves an MMLU accuracy of 68.07\%, which is almost identical to the uniform Int4 baseline (67.95\%). 
For Qwen3-30B-A3B, the robustness is even more pronounced. On GSM8K, the "4/0" strategy at $r=0.9$ achieves 91.74\%, surprisingly surpassing the standard Int4 baseline of 89.08\%. This counter-intuitive gain suggests that the selective application of low-precision quantization to less critical experts may act as a form of regularization.

\noindent\textbf{Impact of Dynamic Quantization (4/2) vs. (4/0).}
Comparing the two dynamic strategies reveals distinct behaviors that depend heavily on model redundancy. For Mixtral-8$\times$7B, the ``4/2'' strategy serves as a vital safety net, particularly under aggressive retention ratios ($r=0.75$). On the CMMLU benchmark, replacing the ``4/0''  strategy with Int2 quantization recovers accuracy from 0.482 to 0.494. This recovery suggests that in Mixtral, non-critical experts still retain significant residual knowledge, and allocating a minimal 2-bit representation is highly effective in preserving model performance with negligible information loss.

\indent Conversely, Qwen3-30B-A3B exhibits higher intrinsic redundancy. In this case, the ``4/0''  strategy often matches or even marginally outperforms the ``4/2'' approach; for instance, achieving 0.9151 versus 0.9090 on GSM8K at $r=0.75$. This implies that for Qwen3-30B-A3B, the distinct functional separation of expert roles allows DyMoE to safely bypass non-critical experts entirely.

\begin{figure}[t]
\centering
\begin{subfigure}[b]{0.48\linewidth}
    \centering
    \includegraphics[width=\linewidth]{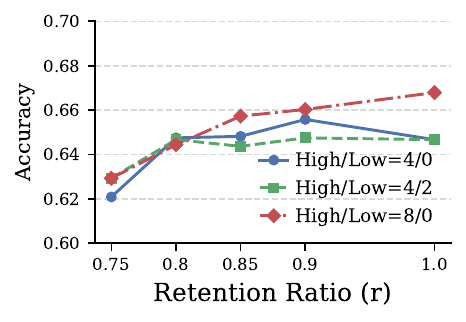}
    \caption{GSM8K}
\end{subfigure}
\hfill
\begin{subfigure}[b]{0.48\linewidth}
    \centering
    \includegraphics[width=\linewidth]{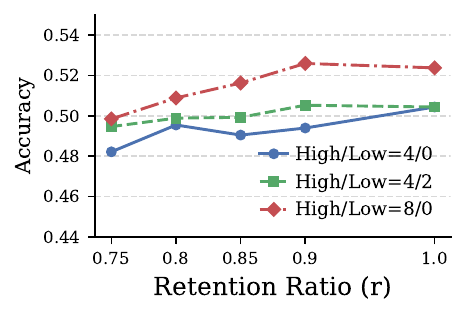}
    \caption{CMMLU}
\end{subfigure}
\caption{Accuracy vs. Retention Ratio. Higher $r$ yields better accuracy, showing DyMoE's flexible trade-off capability.}
\label{fig:DyMoE_retention}
\end{figure}

\begin{table}[t]
\centering
\caption{Evaluation of DyMoE on Mixtral-8$\times$7B and Qwen3-30B-A3B.
The retention ratio $r$ denotes the average proportion of Critical experts preserved across layers.}
\label{tab:combined_results}
\resizebox{\linewidth}{!}{%
\begin{tabular}{@{}llcccc@{}}
\toprule
\textbf{Dataset} & \textbf{Model} & \textbf{High / Low} & \textbf{$r=0.75$} & \textbf{$r=0.9$} & \textbf{$r=1.0$} \\ \midrule
\multirow{4}{*}{\textbf{MMLU}}   & \multirow{2}{*}{Mixtral-8$\times$7B}   & 4 / 0 & 0.6714 & 0.6807 & 0.6795 \\
                                 &                                 & 4 / 2 & 0.6782 & 0.6821 & 0.6795 \\ \cmidrule(l){2-6} 
                                 & \multirow{2}{*}{Qwen3-30B-A3B}  & 4 / 0 & 0.7698 & 0.7737 & 0.7705 \\
                                 &                                 & 4 / 2 & 0.7632 & 0.7704 & 0.7705 \\ \midrule
\multirow{4}{*}{\textbf{CMMLU}}  & \multirow{2}{*}{Mixtral-8$\times$7B}   & 4 / 0 & 0.4822 & 0.4940 & 0.5044 \\
                                 &                                 & 4 / 2 & 0.4947 & 0.5053 & 0.5044 \\ \cmidrule(l){2-6} 
                                 & \multirow{2}{*}{Qwen3-30B-A3B}  & 4 / 0 & 0.7960 & 0.8019 & 0.8044 \\
                                 &                                 & 4 / 2 & 0.7894 & 0.8015 & 0.8044 \\ \midrule
\multirow{4}{*}{\textbf{GSM8K}}  & \multirow{2}{*}{Mixtral-8$\times$7B}   & 4 / 0 & 0.6209 & 0.6558 & 0.6467 \\
                                 &                                 & 4 / 2 & 0.6293 & 0.6475 & 0.6467 \\ \cmidrule(l){2-6} 
                                 & \multirow{2}{*}{Qwen3-30B-A3B}  & 4 / 0 & 0.9151 & 0.9174 & 0.8908 \\
                                 &                                 & 4 / 2 & 0.9090 & 0.9052 & 0.8908 \\ \bottomrule
\end{tabular}%
}
\end{table}

\noindent\textbf{Dynamic Accuracy-Resource Trade-off.}
The relationship between the expert retention ratio $r$ and model accuracy reflects a smooth and controllable trade-off, as visualized in~\autoref{fig:DyMoE_retention}. While accuracy on reasoning-heavy tasks like GSM8K is more sensitive to $r$ when it drops to 0.75, it recovers rapidly as $r$ approaches 0.9. This flexibility provides a tunable knob for real-world deployment: users can dynamically adjust the retention ratio to trade a marginal amount of accuracy for significant latency reduction during peak loads, or increase $r$ to prioritize quality during complex reasoning tasks. Unlike static compression methods, DyMoE enables this runtime adaptation without requiring costly model retraining or per-dataset reconfiguration.

\subsection{Ablation Studies}

To analyze the contribution of each component in DyMoE, we conduct an incremental ablation study on Mixtral-8$\times$7B across two memory-constrained configurations (16~GB and 24~GB). The results, summarized in~\autoref{tab:ablation_strategies}, demonstrate the performance gains from both system-level optimizations and dynamic execution strategies.

\textbf{System-level Optimizations (Rows 1--3).}
The vanilla Load on Demand baseline (Row 1) suffers from severe latency due to the massive I/O bottleneck inherent in transferring large-scale MoE weights over the PCIe bus. By introducing the Expert Cache (Row 2), we achieve a $1.88\times$ to $2.20\times$ speedup in TPOT by mitigating redundant PCIe transfers. The integration of Prefetching (Row 3) further optimizes the pipeline by enabling the overlap of computation and weight transfer, yielding an additional $1.13\times$ to $1.15\times$ acceleration compared to the cache-only baseline.

\textbf{Synergy of Dyquant (Row 4).} Row 4 introduces our Dyquant ($4/2$) strategy but intentionally disables prefetching to isolate the algorithmic gain. Compared directly to the cache-only baseline (Row 2), Row 4 achieves superior decoding efficiency, providing a $1.14\times$ speedup in the 16~GB setting and an even more significant $1.60\times$ speedup in the 24~GB setting. This demonstrates that by dynamically selecting critical experts and using low-precision for sub-critical ones, DyMoE reduces the total I/O volume more effectively than uniform compression.

\textbf{Full System Integration (Rows 5--6).} By combining dynamic execution with system-level optimizations, DyMoE achieves a $2.43\times$ to $4.26\times$ speedup over the Load on Demand baseline. This significant performance gain confirms the effectiveness of our approach in enabling efficient MoE inference on resource-constrained edge devices.

\begin{table}[t]
\centering
\caption{Ablation study of DyMoE dynamic strategies. Dyquant denotes our proposed Dynamic Quantization scheme.}

\label{tab:ablation_strategies}
\resizebox{\linewidth}{!}{
\begin{tabular}{l|cc|cc}
\toprule
\multirow{2}{*}{\textbf{Configuration}} & \multicolumn{2}{c|}{\textbf{16~GB}} & \multicolumn{2}{c}{\textbf{24~GB}} \\ \cline{2-5} 
 & \textbf{TTFT(s)}  & \textbf{TPOT(s)}  & \textbf{TTFT(s)}  & \textbf{TPOT(s)}\\ \midrule
1. Load on Demand  & 1.0193 & 0.2795 & 1.0193 & 0.2795 \\ 
2. Cache & 0.5922 & 0.1489 & 0.4893 & 0.1273 \\ 
3. Cache + Prefetch & 0.5159 & 0.1315 & 0.3954 & 0.1108 \\ \midrule
4.  Cache+Dyquant(4/2)  & 0.5431 & 0.1307 & 0.2423 & 0.0796 \\ 
5. Cache+Dyquant(4/2) & \multirow{2}{*}{0.4269} & \multirow{2}{*}{0.1150} & \multirow{2}{*}{0.2121} & \multirow{2}{*}{0.0754} \\ 
\quad +Prefetcher & & & & \\  
6. \textbf{Cache+Dyquant(4/0)} & \multirow{2}{*}{\textbf{0.3921}} & \multirow{2}{*}{\textbf{0.1048}} & \multirow{2}{*}{\textbf{0.1877}} & \multirow{2}{*}{\textbf{0.0656}} \\ 
\quad \textbf{+Prefetcher} & & & & \\ \bottomrule
\end{tabular}
}
\end{table}

\section{Conclusion}
In this paper, we introduce DyMoE, an algorithm-system co-design for high-performance MoE inference on edge devices. By exploiting runtime expert skewness and depth-wise sensitivity, DyMoE orchestrates mixed-precision cache and prefetching engine to break through the I/O bottleneck. Crucially, DyMoE serves as a plug-and-play solution that requires no costly model retraining or calibration overhead. Our evaluation demonstrates that DyMoE delivers up to 22.7$\times$  and $14.58\times$ speedups in TTFT and TPOT, respectively, while maintaining competitive model accuracy.

\bibliography{example_paper}
\bibliographystyle{icml2026}
\end{document}